\documentclass[runningheads]{llncs}
\usepackage[utf8]{inputenc}
\usepackage[linesnumbered,ruled,vlined]{algorithm2e}
\usepackage{amsmath}
\usepackage{amssymb}
\usepackage{tabularx}

\usepackage{graphicx}
\usepackage{multirow}
\usepackage{changepage}
\usepackage{booktabs}
\usepackage{algorithmic}
\usepackage{todonotes}
\usepackage{url}
\usepackage{comment}
\usepackage{subfig}
\usepackage{subcaption}
\usepackage[title]{appendix}

\usepackage{siunitx}
\usepackage{makecell}
\usepackage{nicematrix}
\usepackage{caption}
\usepackage[hidelinks]{hyperref}

\begin{document}

\title{SpinGPT: A Large-Language-Model Approach to Playing Poker Correctly}
\author{
Narada Maugin* \and Tristan Cazenave*
}

\institute{
*LAMSADE, Université Paris Dauphine - PSL, CNRS, Paris, France 
}

\maketitle

\begin{abstract}
The Counterfactual Regret Minimization (CFR) algorithm and its variants have enabled the development of pokerbots capable of beating the best human players in heads-up (1v1) cash games and competing with them in six-player formats. However, CFR’s computational complexity rises exponentially with the number of players. Furthermore, in games with three or more players, following Nash equilibrium no longer guarantees a non-losing outcome. These limitations, along with others, significantly restrict the applicability of CFR to the most popular formats: tournaments. Motivated by the recent success of Large Language Models (LLM) in chess and Diplomacy, we present SpinGPT, the first LLM tailored to Spin \& Go, a popular three-player online poker format. SpinGPT is trained in two stages: (1) Supervised Fine‑Tuning on 320k high‑stakes expert decisions; (2) Reinforcement Learning on 270k solver‑generated hands. Our results show that SpinGPT matches the solver's actions in 78\% of decisions (tolerant accuracy). With a simple deep-stack heuristic, it achieves 13.4 $\pm$ 12.9 BB/100 versus Slumbot in heads-up over 30,000 hands (95\% CI). These results suggest that LLMs could be a new way to deal with multi-player imperfect-information games like poker.

\end{abstract}

\section{Introduction}

Poker is a family of card games, with no-limit Texas hold'em\footnote{Rules: \url{https://en.wikipedia.org/wiki/Texas_hold\%27em}. 
Glossary: \url{https://en.wikipedia.org/wiki/Glossary_of_poker_terms}.} being its most widespread variant. Originating in the United States in the early nineteenth century, poker has attracted tens of millions of players worldwide, with televised poker broadcasts using hole-card cameras that reveal players’ cards to viewers and online platforms accelerating its spread since the early 2000s. Unlike games of complete information such as chess, poker is a game of incomplete information: each player holds private cards unknown to their opponents. Moreover, bet sizes are flexible. As a result, the game tree is immense, ranging from $10^{13}$ nodes in heads-up (HU) limit to over $10^{160}$ in heads-up no-limit, and even more as the number of players increases, making an exhaustive search infeasible \cite{johansonMeasuringSizeLarge2013b}. 

Accordingly, poker has for many decades served as a testbed for game theory and artificial intelligence \cite{ferguson2003borel,findler1977studies}. Substantial progress has been made, leading to programs that compete with—and in some cases surpass—top human players in specific variants. However, most progress has concerned two-player cash games; tournament play differs fundamentally because the number of players and stack sizes change as the game progresses. Furthermore, maximizing chip expected value (cEV) is no longer optimal, as the value of chips is non-linear: doubling one's stack does not double one's chances of winning, rendering the mere accumulation of chips suboptimal\footnote{This non-linearity is often modeled by the ICM (Independent Chip Model). It formalizes the fact that as the tournament progresses, the strategy evolves to place greater emphasis on survival and reaching the money.}. As a consequence, hands played can no longer be considered independent and identically distributed and the algorithms used must be reconsidered. A popular format illustrating these difficulties is the Spin \& Go, a hyper-turbo three-player mini-tournament. Although simpler than a large tournament, this format already introduces the dynamics of varying player counts and stack sizes.

\section{Related Work}

Before regret-minimization methods, computer poker primarily relied on knowledge-based rules, heuristic evaluation, and early equilibrium-driven abstractions. These approaches achieved competent play in constrained variants but struggled with the full game’s combinatorial complexity, conceding they were “nowhere close to solving full-scale poker” \cite{billingsChallengePoker2002}.

A major breakthrough came with the introduction of the Counterfactual Regret Minimization (CFR) algorithm in 2007 \cite{zinkevichRegretMinimizationGames2007}. This algorithm and its subsequent variations allow an agent to approximate an optimal strategy by learning to minimize the regret associated with suboptimal decisions. In 2015, the Cepheus program computed a near-perfect Nash equilibrium strategy for heads-up limit hold'em, thereby weakly solving this variant \cite{bowlingHeadsupLimitHoldem2015}. Two years later, in the far more complex no-limit variant, two independent programs achieved significant results. DeepStack \cite{moravcikDeepStack2017} defeated professional players in an online match, and shortly after, Libratus \cite{brownLibratus2018} surpassed some of the world's top professionals in a live casino match. These programs relied on state-space abstractions (such as treating the ace and king of hearts similarly to the ace and king of spades pre-flop) and real-time subgame solving techniques based on improved versions of CFR.

Following these successes in one-on-one settings, researchers turned to the harder problem of multi-player poker. This step was taken in 2019 with Pluribus \cite{brownPluribus2019}, which defeated professionals in six-player no-limit Texas hold'em, though one top professional finished roughly even. Although its strategy computation required substantial resources, this was the first AI to surpass human experts in a multi-player setting. To overcome the remaining computational difficulties, new approaches have emerged \cite{brownRebel2020a,zhaoAlphaHoldemHighPerformanceArtificial2022a}, but without exceeding the performance of the CFR paradigm.

Using a large language model for poker might appear counter-intuitive, because even frontier models have proven incapable of playing competently through direct prompting (zero-shot or few-shot): their pre-training corpora contain very few high-quality hand histories compared with games such as chess. This is why competitive LLM play in chess is viable, as evidenced by the tournament organized by Kaggle and Google \cite{KaggleChess2025}. However, fine-tuning has proved effective at closing this gap \cite{zhuangPokerBench2025}. PokerGPT \cite{huangPokerGPT2024} followed this route by fine-tuning an LLM, but it was limited by its training data, which consisted exclusively of spectator data from low-stakes games (NL5). This meant that players' private cards were often unknown, and the dataset was biased by the flawed strategies of weak players. Our work directly addresses these shortcomings.

\section{Methodology: A Large-Language-Model Approach}

Our hypothesis is that a properly trained LLM can play poker effectively, owing to (i) broad prior knowledge (rules, basic strategy, probabilities) and (ii) facility with learning the syntax and semantics of structured representations. 

Our methodology follows a two-stage training process: first, a Supervised Fine-Tuning (SFT) phase to teach the model the language and the strategy of a professional poker player, followed by a Reinforcement Learning (RL) phase using ORPO (Optimistic Model Rollouts for Pessimistic Offline Policy Optimization) algorithm \cite{zhai2024orpo} on solver-generated hands to refine its strategy toward game-theoretic optimality.

\subsection{Model Architecture}
SpinGPT is based on Llama-3.1-8B-Instruct (Meta, Llama 3.1 Community License), an 8-billion parameter open-weight language model from Meta. This model represents a trade-off between representational capacity—necessary for modeling complex poker strategies—and computational requirements, making it tractable on accessible hardware infrastructure. This model demonstrated performance comparable to or superior to that of other models of similar size \cite{grattafiori2024llama3}, leading us to believe in the potential of the 8B model to achieve state-of-the-art results for our task. We also conducted tests with a smaller model, Llama-3.2-1B-Instruct. 

\subsection{Data and Preprocessing}
We constructed a dataset comprising 320,000 poker decisions from 8,800 Spin \& Go games. This data was sourced from one of this paper's authors, Narada Maugin, and corresponds to hands he played professionally on the PartyPoker.fr platform between 2018 and 2020. The games were played at high-stakes buy-ins of \texteuro 50, \texteuro 100, and \texteuro 250. 

The raw hand history files have been anonymized. Because they contain superfluous information such as timestamps and table chat, we parsed them to retain only strategically relevant fields (stacks, positions, actions, public and private cards). Throughout, we express all stack sizes in big blinds (BB).
We then transformed each decision point into a structured, text-based “instruction” prompt that the LLM can read. For instance, a typical game situation is encoded as the following instruction: \texttt{"pos:H=BTN stacks:\allowbreak H=29.3,\allowbreak BB=1.7,\allowbreak SB=19.0 hand:TsQs | pre:H r2,SB c,BB f | flop:4h7s6c SB b1,\allowbreak H c |\allowbreak  turn:\allowbreak 8d SB b1,\allowbreak H c | river:\allowbreak 9c SB b1 H:"}. This structured encoding contains all essential information. Indeed, in this example, the following are present: player positions (Hero denotes the focal player whose private cards and decisions are logged. Here, Hero is on the button), stack depths (29.3 BB for Hero, 1.7 BB for the big blind player, and 19.0 BB for the Small Blind player), Hero's cards (TsQs, meaning Ten and Queen of spades), and the sequence of actions on each street. The model, receiving this description as input, must provide the appropriate action as output (r6.5 in the example, meaning raise to 6.5 BB). The set of all possible actions was represented by a compact vocabulary (c for call, f for fold, x for check, a for all-in, r\{amount\} and b\{amount\} for raise and bet of a certain amount, respectively), which was integrated into the model's tokenizer. Importantly, the model is not restricted to a predefined set of actions and bet sizes; it can generate any tokens and therefore any numerical value for \{amount\}.

\subsection{Supervised Fine-Tuning}

Then, we fine-tuned the model on our poker dataset using standard supervised learning via the LLaMA-Factory framework \cite{Zheng_LlamaFactory_Unified_Efficient_2024}. We employed Low-Rank Adaptation (LoRA) \cite{hu2021lora} to drastically reduce training costs. This method involves adding a few trainable parameters (approximately 10 million in our case) to modulate the original weights. Our main hyperparameters were as follows: each LoRA adapter used rank $r=8$ and scaling factor $\alpha = 16$ (no dropout, no additional bias weights). Training ran for four epochs on sequences of up to 128 tokens, starting from a learning rate of $5 \times 10^{-5}$ that decayed with a cosine schedule. The entire fine-tuning took 10 hours on a single NVIDIA A100 (40 GB) in FP16 precision. We refer to this model as SpinGPT-SFT.

\subsection{Offline Reinforcement Learning}

While effective, the model produced by SFT (SpinGPT-SFT) merely replicates a specific human strategy from several years ago, inheriting its potential biases and weaknesses. To transcend imitation and move toward an optimal strategy, a second training phase was implemented to align the model with a Game Theory Optimal (GTO) Nash equilibrium.

To achieve this, we chose an offline RL approach. We leveraged an external, CFR-based GTO solver named InstaGTO to provide near-optimal decisions. Therefore, we used InstaGTO to create a dataset of 270,000 synthetic hands, covering a wide array of post-flop situations (1/3 are HU, 1/3 are SBvBB in 3-way, 1/6 are SBvBTN in 3-way, and 1/6 are BBvBTN in 3-way). A significant challenge with solver-generated data is its sterile, perfect nature, which differs from real-world play (e.g., it lacks pre-flop action, only covers two-player post-flop scenarios, involves specific stack depths, and contains no idiosyncratic human errors). Training exclusively on this data could lead to catastrophic forgetting of the robust patterns learned during SFT. To mitigate this, we created a mixed dataset by retaining 50,000 hands from the original human dataset.

This combined dataset of 320,000 hands was used to refine the model with the ORPO (Optimistic Model Rollouts for Pessimistic Offline Policy Optimization) algorithm \cite{zhai2024orpo}. ORPO is suited for this task as it combines SFT with preference alignment. Rather than simply maximizing a reward signal like Expected Value (EV), ORPO adjusts the model's policy to increase the likelihood of the solver's action (we used the solver’s argmax-EV action as the positive label) relative to other suboptimal actions.

This final stage, run also with LLaMA-Factory, added 10 GPU-hours on an NVIDIA A100. It reused mainly the SFT LoRA configuration and applied QLoRA to the base model (4-bit NF4 with double quantization). We optimized with ORPO ($\beta=0.1$) for two epochs on sequences up to 128 tokens, using a learning rate of $2 \times 10^{-5}$; the optimizer ran in BF16. The model resulting from this two-stage training pipeline is our final agent, SpinGPT. 

\section{Experimental Results}

To evaluate our models, we use two test sets that were held out during training: a Professional test set, which contains 32,000 decisions from real hands played by the human expert, and a Solver test set, which includes 30,000 decisions generated by the InstaGTO solver across various post-flop configurations.

\subsection{Evaluation Metrics}
We first measure performance using two types of accuracy. Exact Accuracy is the percentage of predictions that perfectly match the ground-truth action, including both the action type and the exact amount (e.g., 'r4.6'). As a more lenient alternative, we also report Tolerant Accuracy, which considers a prediction correct if the action type matches and the bet or raise amount is within a $\pm 0.5$ BB tolerance of the true value. For instance, if the model predicts a bet of 1.5 BB (b1.5) while the true action was a bet of 1.6 BB (b1.6), the prediction is counted as correct under this metric.

To measure how well the model predicts each poker action class (\textit{bet}, \textit{call}, \textit{fold}...), we report the macro $F_1$ score. For every action class $c$, we first compute an $F_1$ score: the harmonic mean of precision (the fraction of predicted actions of class $c$ that are correct) and recall (the fraction of true actions of class $c$ that are recovered). We then take the mean of these per-class scores to obtain the macro average. 

Finally, neither accuracies nor the macro $F_{1}$ score evaluates the quantitative correctness of bet sizing. For instance, if the expected move is bet 1 BB (b1) but our model gives bet 5 BB (b5), it falls into the ‘bet’ class but is quite far from the expected value. So, to evaluate the accuracy of bet and raise amounts, we use two other metrics. The mean absolute error (MAE), expressed in BB, and the mean absolute percentage error (MAPE), expressed as a percentage. A lower MAE and MAPE indicate better calibration of bet sizes.

\subsection{Imitation performance of the SFT Model}

The initial model, SpinGPT-SFT, was first evaluated on its ability to imitate the professional player. As shown in Table 1, it achieves an exact accuracy of 80\%. Its macro $F_{1}$ score of 86\% indicates an ability to classify all types of actions correctly. For comparison, PokerGPT, another LLM-based agent, achieved a macro $F_{1}$ score of 78\% on human player data.

It is critical to note that a 100\% score is theoretically unattainable. An expert's strategy is inherently stochastic; to remain unpredictable, a player employs mixed strategies, such as raising with a certain hand 70\% of the time and calling 30\% of the time in the exact same situation, and adapts to their opponent, playing differently against an amateur or professional player. This is compounded by noise from human errors.

% Début matrice de confusion
\begin{figure}[ht]
\centering
\caption{Confusion matrix of SpinGPT-SFT on the Professional test set.}
\label{fig:conf_matrix}

\sffamily

\begin{NiceTabular}{c c | p{1.5cm} | p{1.5cm} | p{1.5cm} | p{1.5cm} | p{1.5cm}}[
    cell-space-limits = 4pt,
    ]

& & \Block{1-1}{\bfseries bet} & \Block{1-1}{\bfseries call} & \Block{1-1}{\bfseries check} & \Block{1-1}{\bfseries fold} & \Block{1-1}{\bfseries raise} \\
\Hline

\Block[borders={}]{5-1}{\rotatebox{90}{\bfseries True Label}}
& \bfseries bet   
& \cellcolor{blue!75} \Block{1-1}{\color{white} 2659 \\ \small(74.6\%)}
& \cellcolor{blue!0} \Block{1-1}{\color{black} 0 \\ \small(0\%)}
& \cellcolor{blue!30} \Block{1-1}{\color{black} 874 \\ \small(24.5\%)}
& \cellcolor{blue!0} \Block{1-1}{\color{black} 0 \\ \small(0\%)}
& \cellcolor{blue!1} \Block{1-1}{\color{black} 31 \\ \small(0.9\%)} \\
\Hline

& \bfseries call  
& \cellcolor{blue!10} \Block{1-1}{\color{black} 0 \\ \small(0\%)}
& \cellcolor{blue!86} \Block{1-1}{\color{white} 5556 \\ \small(85.8\%)}
& \cellcolor{blue!10} \Block{1-1}{\color{black} 0 \\ \small(0\%)}
& \cellcolor{blue!15} \Block{1-1}{\color{black} 500 \\ \small(7.7\%)}
& \cellcolor{blue!14} \Block{1-1}{\color{black} 422 \\ \small(6.5\%)} \\
\Hline

& \bfseries check 
& \cellcolor{blue!19} \Block{1-1}{\color{black} 743 \\ \small(9.4\%)}
& \cellcolor{blue!0} \Block{1-1}{\color{black} 0 \\ \small(0\%)}
& \cellcolor{blue!89} \Block{1-1}{\color{white} 7009 \\ \small(88.9\%)}
& \cellcolor{blue!0} \Block{1-1}{\color{black} 1 \\ \small(0\%)}
& \cellcolor{blue!3} \Block{1-1}{\color{black} 135 \\ \small(1.7\%)} \\
\Hline

& \bfseries fold  
& \cellcolor{blue!0} \Block{1-1}{\color{black} 0 \\ \small(0\%)}
& \cellcolor{blue!9} \Block{1-1}{\color{black} 403 \\ \small(4.6\%)}
& \cellcolor{blue!0} \Block{1-1}{\color{black} 0 \\ \small(0.0\%)}
& \cellcolor{blue!94} \Block{1-1}{\color{white} 8141 \\ \small(93.9\%)}
& \cellcolor{blue!2} \Block{1-1}{\color{black} 124 \\ \small(1.4\%)} \\
\Hline

& \bfseries raise 
& \cellcolor{blue!1} \Block{1-1}{\color{black} 37 \\ \small(0.6\%)}
& \cellcolor{blue!20} \Block{1-1}{\color{black} 579 \\ \small(9.8\%)}
& \cellcolor{blue!7} \Block{1-1}{\color{black} 210 \\ \small(3.5\%)}
& \cellcolor{blue!6} \Block{1-1}{\color{black} 184 \\ \small(3.1\%)}
& \cellcolor{blue!83} \Block{1-1}{\color{white} 4926 \\ \small(83.0\%)} \\
\Hline

& & \Block[borders={}]{1-5}{\bfseries Predicted Label} & & & & \\

\end{NiceTabular}
\end{figure}
% Fin matrice de confusion

The confusion matrix (Figure 1) shows high action-match, correctly identifying actions most of the time, especially when folding (94\% of cases). More importantly, the model produces a very small number of illegal moves (1.4\% overall). For example, in 135 instances where the correct action was check, the model predicted raise. A raise is only legal in response to a prior bet, so this prediction is impossible. This suggests the model has confused the token for a bet with the token for a raise. These types of errors, which account for a small fraction of predictions, can be easily corrected in deployment by mapping the invalid prediction to its closest legal semantic equivalent (e.g., mapping a predicted raise to a bet, or a predicted call to a check, when no prior bet exists).

\begin{table}[ht]
    \centering
    \caption{Comparison of the performance of the versions on the Professional and Solver test set.}
    \label{tab:full_performance}
    
    % en-têtes en gras
    \renewcommand\theadfont{\bfseries}

    % 'tabularx' s'adapte à la largeur de la page
    \begin{tabularx}{\textwidth}{
        ll      % Modèle, Dataset
        S[table-format=2.1] % Précision Exacte
        S[table-format=2.1] % Précision Tolérante
        S[table-format=2.1] % Coup Illégal
        S[table-format=1.2] % F1-Score
        S[table-format=1.2] % MAE (BB)
        S[table-format=2.1] % MAPE (%)
    }
        \toprule
        % \thead (de makecell) permet des en-têtes sur plusieurs lignes
        \thead{Model} & \thead{Dataset} & {\thead{Exact\\Accuracy (\%)}} & {\thead{Tolerant\\Accuracy (\%)}}  & {\thead{Macro $F_{1}$ \\ (\%)}} & {\thead{MAE\\(BB)}} & {\thead{MAPE\\(\%)}} \\
        \midrule

        SpinGPT-SFT  & Professional & 80 & 84 & 86 & 0.61 & 19 \\
        SpinGPT & Professional & 79 & 83 & 84 & 0.30 & 12 \\
        SpinGPT & Solver       & 72 & 78 & 77 & 0.93 & 35 \\
        
        \bottomrule
    \end{tabularx}
    
\end{table}

\subsection{Performance of the final SpinGPT model}

After reinforcement learning, the final SpinGPT model shows slightly different results. On the Professional test set, its imitation accuracy drops to 79\%, and its macro $F_{1}$ score to 83\% (table 1). This small decrease does not indicate a regression, but rather suggests that the model is no longer simply copying human actions—it has started to adjust its behavior based on solver feedback. The MAE improved from 0.61 BB to 0.30 BB, a reduction that yields predictions within one-third of the minimum bet (1 BB).

On the Solver test set, performance is lower across all metrics: macro $F_{1}=77\%$, exact accuracy $=72\%$, tolerant accuracy $=78\%$, $\mathrm{MAE}=0.93$ and $\mathrm{MAPE}=0.35$. This is expected, as GTO policies rely on finely balanced mixed strategies and precise bet sizes, which are difficult to understand and implement. 

\subsection{Gameplay Performance Evaluation}

While imitation metrics confirm the model's ability to learn a strategy, they do not measure its effectiveness in actual gameplay. To assess this, we conducted two head-to-head confrontations.

\subsubsection{Confrontation with a Benchmark Agent.}

We benchmarked SpinGPT-SFT against Slumbot \cite{jacksonSlumbot2013}, the 2018 AAAI Computer Poker Competition champion. Slumbot is designed for heads-up cash games with a constant 200 BB stack, whereas SpinGPT-SFT targets Spin \& Go tournaments that begin 3-handed at 25 BB and transition to HU as stacks evolve. Despite this mismatch, Slumbot’s public availability, well-documented API, and published results make it the strongest standardized reference—no high-performance bot exists for the 3-player tournament format to our knowledge.

The supervised model initially lost heavily because it shoved all-in at 200 BB far too often—an action that is correct at 25 BB and less but catastrophic at deep stacks. This failure mode reflects the absence of 200 BB examples in the training data. To obtain a fair baseline we applied a minimal heuristic: each all-in bet was replaced by a 2/3 pot bet if the agent was the first to bet, or a raise to 3x the opponent's bet otherwise. This preserves the aggressive intent of the model while avoiding losing the entire stack from time to time.

With this patch, over 30,000 hands, SpinGPT-SFT achieved a win rate of 13.4 $\pm$ 12.9 BB/100 (95\% CI)\footnote{The BB/100 metric is the most commonly used in poker. It represents the number of big blinds won per 100 hands on average. The basic strategy, which consists of always folding, is -75 BB/100. Two equivalent players will be at 0 BB/100, and one player is clearly winning against another starting at 6 BB/100.}. Table 2 situates this result among other state-of-the-art research agents. Results from prior work are cited from the original papers; confidence interval (CI) computation methods may differ across papers.

\begin{table}[ht]
    \centering
    \caption{Comparison of win rates against the Slumbot benchmark agent.}
    \label{tab:slumbot_comparison}
    \sisetup{table-format=2.2, table-align-uncertainty=true}
    \begin{tabular}{@{}l S[table-format=2.2] l c c@{}}
        \toprule
        \textbf{Agent} & {\thead{\textbf{Win Rate}\\(BB/100)}} & {\thead{\textbf{95\% CI}\\(BB/100)}} & \textbf{Hands Played} & \textbf{Year} \\ \midrule
        Baby Tartanian8 \cite{brownBabyTartanian8Winning2016} & 3.6 & $\pm$ 1.2 & N/A & 2016 \\
        ReBeL \cite{brownRebel2020a} & 4.5 & $\pm$ 1.0 & N/A & 2020 \\
        AlphaHoldem \cite{zhaoAlphaHoldemHighPerformanceArtificial2022a} & 11.2 & $\pm$ 1.6 & \num{100000} & 2022 \\
        PokerGPT \cite{huangPokerGPT2024} & 15.8 & $\pm$ 4.9 & \num{10000} & 2024 \\
        SpinGPT-SFT (Ours) & 13.4 & $\pm$ 12.9 & \num{30000} & 2025 \\ \bottomrule
    \end{tabular}
\end{table}

\subsubsection{Confrontation between our two agents.}

To quantify the improvement provided by the reinforcement learning phase, we conducted a head-to-head match between the final SpinGPT model and the SFT-only model (SpinGPT-SFT). The match was played over 10,000 hands in a duplicate format (where players switch cards after each hand to reduce variance) at a 25 BB stack depth. The final SpinGPT model achieved a win rate of 13.2 $\pm$ 7.24 BB/100 against the SFT-only version. This result shows that the reinforcement learning phase brings about a very significant improvement in game performance.

\section{Conclusion and Discussion}
In this work, we introduced SpinGPT, an LLM-based agent for the Spin \& Go poker format. We employed a two-stage training pipeline combining Supervised Fine-Tuning on professional hand histories with Reinforcement Learning using a GTO solver. 

The SFT variant (SpinGPT-SFT) is trained on 320k decisions and outperformed the benchmark pokerbot Slumbot after just one heuristic all-in patch. This is notable given its deep-stack errors like limping—reasonable at $\le 25$\ BB but a mistake at 200 BB. We then applied RL with ORPO to 270k solver-generated hands. The final agent, SpinGPT, achieves 78\% tolerant accuracy on solver actions and clearly beats SpinGPT-SFT.

However, our work has limitations. First, SpinGPT has not yet faced strong human opposition in tournament play; this is the primary route to establish SpinGPT’s true level. Future tests should instead target human opponents or bots specialized in Spin \& Go (or at least heads-up with shallow stacks) to better gauge SpinGPT’s actual strength. We relied on imitation, but it is difficult to know what this represents in real play. Second, the final model has not been evaluated against Slumbot because large-scale matches are costly and the benchmark format does not align. Third, as an LLM, SpinGPT can exhibit typical language-model errors and hallucinations—we observed numerical mistakes such as treating “5.11” as greater than “5.2”.

\subsubsection{Ethical and environmental considerations} To reduce misuse risk, we publicly release neither the model weights nor the full training data; unrestricted release could enable terms-of-service-violating poker bots on real-money platforms. However, we provide access to these materials to researchers upon request. On the environmental side, counting both training and evaluation games, our total compute footprint is estimated at 5.2 kgCO2e using Green Algorithms \cite{lannelongueGreenAlgorithmsQuantifying2021}. 

\subsubsection{Future work}
Future work could advance on several fronts. Immediate improvements include the use of more powerful foundation models and even higher-quality hands. In addition, a Retrieval-Augmented Generation (RAG) module could query pre-computed GTO charts on demand, ensuring optimal preflop and push-or-fold actions. To enable deeper reasoning, computationally heavier techniques such as chain-of-thought prompting could also be explored. Adaptivity is another key objective: training on data labeled by opponent skill—or letting the LLM infer tendencies from hand history—could teach the agent to adjust its style and exploit weaknesses. Beyond raw performance, explainability is another avenue. When trained on curated human data, an LLM can behave in a human-like, intuitive manner. It could therefore generate natural-language explanations of strategic choices and opponent tendencies—useful for match commentary and personalized coaching. Finally, coupling the LLM with a solver (that tackles only heads-up post-flop) supplies a GTO baseline, while the LLM governs every other spot—and even chooses to deviate from that baseline whenever exploitation seems profitable. By pursuing these directions, it may be possible for LLM-based agents to reach expert-level performance in tournaments, suggesting the potential of language models to tackle complex, imperfect-information games.

\bibliographystyle{plain}
\bibliography{main}

\end{document}